\documentclass[10pt,twocolumn,letterpaper]{article}

\usepackage{wacv}
\usepackage{times}
\usepackage{epsfig}
\usepackage{graphicx}
\usepackage{amsmath}
\usepackage{amssymb}
\usepackage{mathtools} 
\usepackage{tabularx}
\usepackage{booktabs}
\usepackage[utf8]{inputenc}

\usepackage[pagebackref=true,breaklinks=true,letterpaper=true,colorlinks,bookmarks=false]{hyperref}
\makeatletter
\def\and{%
  \end{tabular}%
  \hskip 0.00em \@plus.01fil\relax
  \begin{tabular}[t]{c}}
\makeatother

  \wacvfinalcopy 

\def\wacvPaperID{7} 

\ifwacvfinal\fi
\title{Automatic 3D Reconstruction of Manifold Meshes\\via Delaunay Triangulation and Mesh Sweeping}

\author{Andrea Romanoni\\
Politecnico di Milano\\
Italy\\
{\tt\scriptsize andrea.romanoni@polimi.it}
\and
Amaël Delaunoy\\
ETH Zürich \\
Switzerland\\
{\tt\scriptsize amael.delaunoy@gmail.com}
\and
Marc Pollefeys\\ 
ETH Zürich \\
Switzerland\\
{\tt\scriptsize marc.pollefeys@inf.ethz.ch}
\and
Matteo Matteucci\\
Politecnico di Milano\\
Italy\\
{\tt\scriptsize matteo.matteucci@polimi.it}
}

\begin{document}

\maketitle
\ifwacvfinal\thispagestyle{empty}\fi

\begin{abstract}
In this paper we propose a new approach to incrementally initialize a manifold surface for automatic 3D reconstruction from images.
More precisely we focus on the automatic initialization of a 3D  mesh as close as possible to the final solution; indeed many approaches require a good initial solution for further refinement via multi-view stereo techniques. 
Our novel algorithm automatically estimates an initial manifold mesh for surface evolving multi-view stereo algorithms, where the manifold property needs to be enforced. It bootstraps from 3D points extracted via Structure from Motion, then iterates between a state-of-the-art manifold reconstruction step and a novel mesh sweeping algorithm that looks for new 3D points in the neighborhood of the reconstructed manifold to be added in the manifold reconstruction.
The experimental results show quantitatively that the mesh sweeping improves the resolution and the accuracy of the manifold reconstruction, allowing a better convergence of state-of-the-art surface evolution multi-view stereo algorithms.
\end{abstract}

\section{Introduction}
\label{sec:intro}
Reconstructing the observed scene from a set of images is a widely studied problems in computer vision: it is known as \emph{Structure from Motion} (SfM), when the aim is to reconstruct both the scene (the structure) and the camera poses (the motion), or as \emph{multi-view stereo} (MVS), when camera poses are known.
While SfM techniques provide a sparse reconstruction of the environment, MVS aims at reconstructing a dense and accurate model of the observed scene. 

Since the datasets of \cite{Seitz_et_al06} and \cite{strecha2008} were made available, dense MVS has been faced with different approaches, some of which reaching very accurate results; however, some issues are still open, e.g., the initialization and the management of untextured regions and the handling of false matches.
According to the representation of the 3D model, MVS methods can be subdivided in \emph{point-based}, \emph{volumetric} and \emph{mesh-based} methods.
Point-based methods estimate the depth of each image pixel, reconstructing a point cloud \cite{Fu10,Tola12}; they provide accurate and dense point clouds, but the outcome is often redundant where the surface is flat, and they are not able to reconstruct untextured area.
Volumetric methods, first proposed by \cite{curless1996volumetric}, build the reconstruction by discretizing the space and labeling as matter or free space the voxels, as in \cite{curless1996volumetric}, or the tetrahedra, as in \cite{labatut2007efficient}.
Voxel-based methods  require a huge amount of memory, and even if attempts to make the representation more compact exist \cite{steinbrucker2014volumetric}, they are still not suitable for large-scale reconstruction.
Tetrahedron-based methods are more compact, but the accuracy of the reconstruction still needs to be refined to reach state-of-the-art results.
This latter approach is often used as an initialization to mesh-based methods in \cite{vu_et_al_2012,hiep2009towards,salman2010surface}.

Mesh-based methods have been proven to be suitable to build continuous, high accurate reconstructions of both small objects and large-scale scenes \cite{hiep2009towards,vu_et_al_2012,salman2010surface}.
These methods refine an existing mesh by minimizing an image similarity measure such as the Zero Mean Cross Correlation (ZNCC) \cite{hiep2009towards,pons2007multi,zaharescu2007transformesh} or the Sum of Squared Differences (SSD) \cite{Delaunoy_et_al_08,delaunoy2011gradient}. 
The initialization of the existing mesh is one of the major issues in state-of-the-art mesh-based MVS methods. For single object reconstruction, as in the Middelbury dataset \cite{Seitz_et_al06}, the initial mesh is usually estimated by the visual hull \cite{laurentini1994visual}, but this is not applicable in more complex scene such as the ones in \cite{strecha2008} or in large-scale scenarios.

One of the most effective and scalable mesh-based algorithm for multi-view stereo was proposed by Vu \etal \cite{vu_et_al_2012}; in their work, the authors initialize the mesh evolution algorithm by extracting a very dense and noisy point cloud, then building a Delaunay triangulation out of it, and estimating the initial mesh via a \emph{s-t} cut algorithm based on the work of \cite{labatut2007efficient}. 
This last work inspired other extensions such as \cite{jancosek2011multi}.
The main problem with this approach is that the \emph{s-t} cut does not guarantee the output mesh to be a manifold while the \emph{surface evolution} algorithm needs this property; indeed Vu \etal  need to manually check the outcome of the \emph{s-t} cut before surface evolution.

Some mesh-based algorithms, e.g., \cite{pan2015automatic,li2015detail}, initialize the reconstruction with a point-based MVS such as CMVS \cite{Fu10} together with Poisson Reconstruction \cite{kazhdan2006poisson}.
This process automatically provides an initialization usually close to be manifold, but still without guarantees, and it has issues related to redundancy of the points and non-scalability.
In the literature, some methods  to estimate a manifold mesh exist, but most of them rely on silhouettes \cite{vogiatzis2005multi,furukawa2006carved,esteban2004silhouette}, thus they are limited to small objects, or they are not scalable due to the usage of voxel-based reconstruction \cite{hornung2006hierarchical}.
A suitable approach to estimate a manifold mesh has been recently proposed in \cite{lhuillier_Yu2013,Litvinov_Lhuillier_13,Romanoni15a,Romanoni15b}: the authors incrementally reconstruct the scene from very sparse data which are the outcome of SfM algorithms. 
For many applications the reconstruction relying only on these points is sufficient, but for a surface evolution approach, it needs resolution and accuracy improvements. Indeed, to avoid getting stuck in local minima during the evolution process, also Vu \etal need to densify the point cloud extracted by the SfM algorithm.

In this paper we propose a novel and fully automatic approach to estimate a manifold, suitable for being a good initialization for a surface evolution mesh-based MVS algorithm.
We bootstrap from a manifold reconstructed in a similar fashion to \cite{Litvinov_Lhuillier_13}, and we refine it by iteratively sweeping the triangles of the manifold around their neighborhood looking for good stereo matches.
This approach takes inspiration from the multi-directions plane-sweep of \cite{gallup2007real}, where the authors sweep a set of planes choosing fixed directions related to the direction of the walls identified from Structure from Motion points; for each pixel and each plane they collect the matching costs among the views and choose the best stereo matching cost to estimate depth maps.
Another interesting method that performs local plane sweeping is presented in \cite{sinha2014efficient}, where, again, the output is a depth map.
In the proposed approach we estimate directly a manifold mesh; the directions of the planes are automatically driven by the data, and we avoid to sweep on the whole scene, since we look for new 3D points in the neighborhood of the iteratively reconstructed manifold. 
Differently from \cite{gallup2007real}, we directly recover a consistent dense 3D mesh instead of independent depth maps.


In Section \ref{sec:manifold} we summarize the reconstruction and refinement algorithm that builds incrementally a manifold and refines it. In Section \ref{sec:experimental} we show the results of our approach, underlying the improvement of the reconstruction accuracy thanks to the sweeping algorithm on both simulated and real datasets. In Section \ref{sec:conclusion} we conclude pointing out possible future work.

\begin{figure}[t]
\centering
\begin{tabular}{ccc}
\includegraphics[width=0.28\columnwidth]{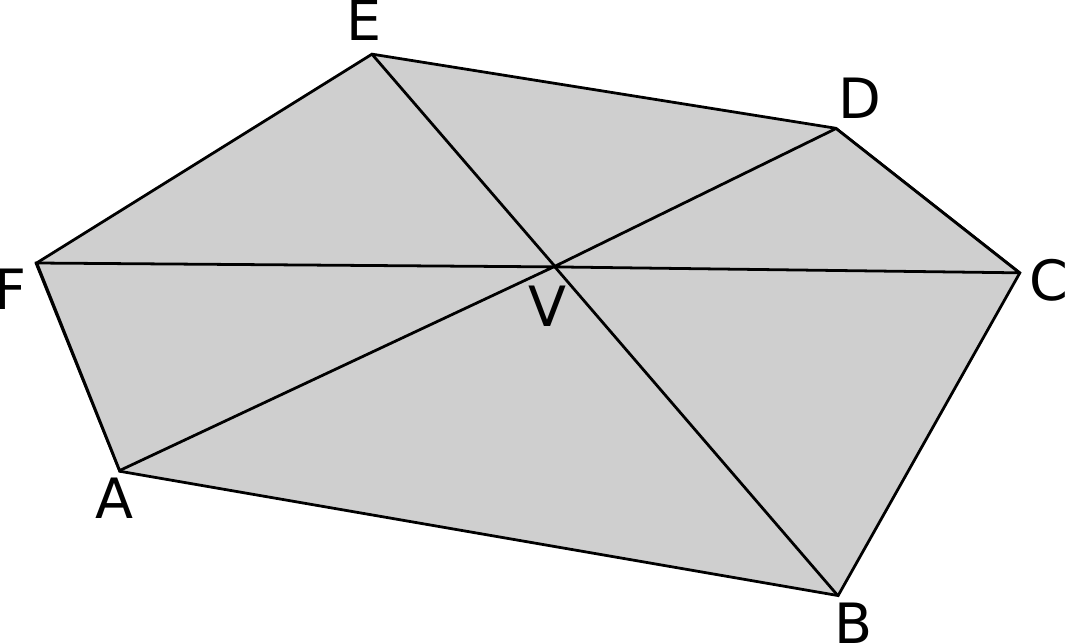}&
\includegraphics[width=0.28\columnwidth]{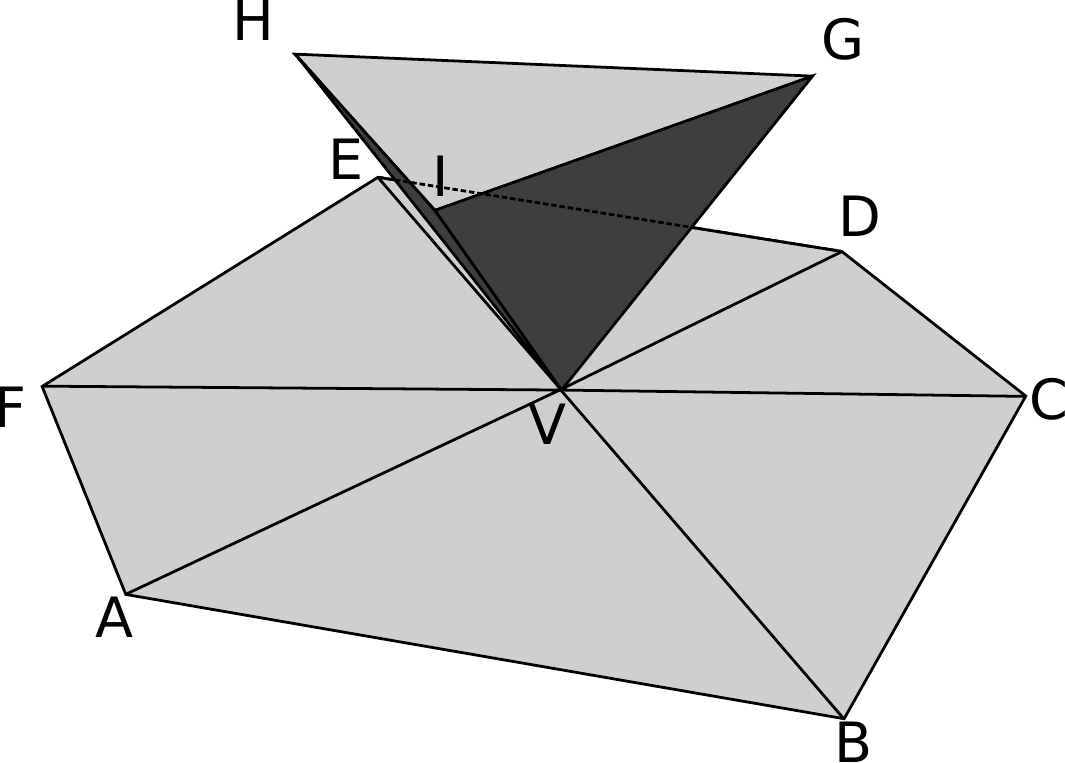}&
\includegraphics[width=0.28\columnwidth]{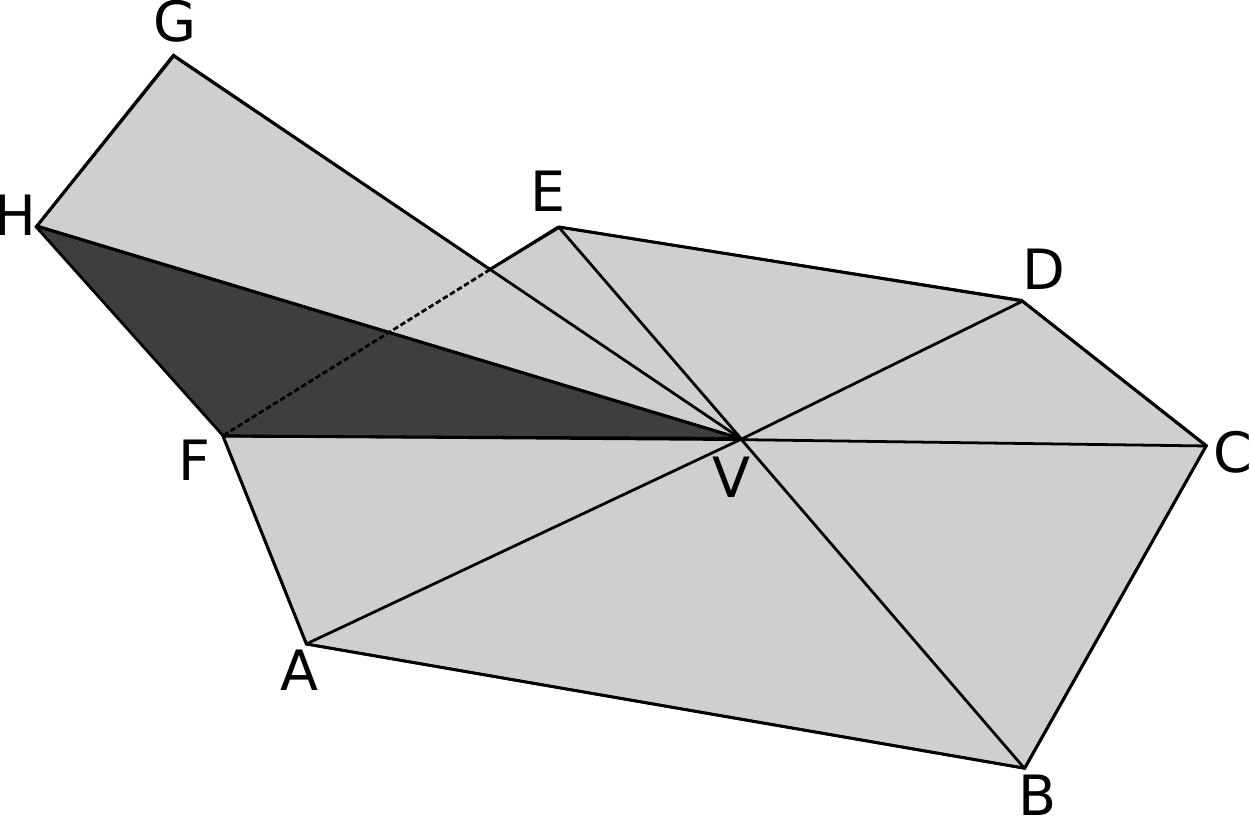}\\
(a)&(b)&(c)
\end{tabular}
\caption{The manifold property: the vertex $V$ in (a) is regular ($ABCDEF$ is a closed path without cycles), while in (b) and (c) the vertex $V$ is not regular since, in the former case the path $ABCDEFGHI$ is not closed, and in the latter $ABCDEFGHF$ has two cycles.}
\label{fig:vertexManifold}
\end{figure}

\section{Proposed algorithm}
\label{sec:manifold}
In the proposed system, we iteratively alternate manifold reconstruction and mesh sweeping. 
First we describe how we estimate the initial mesh, then how we extract new points through mesh sweeping and how we add them, incrementally, to the manifold.
\subsection{Manifold Reconstruction}
\label{subsec:manifold}
Let recall that the manifold property holds if and only if the neighborhood of each surface point is homeomorphic to a disk. 
So a mesh, which is a discrete surface, is manifold if and only if each vertex $v$ is \emph{regular}, i.e., if and only if the edges opposite to $v$ form a closed path without loops  \cite{Litvinov_Lhuillier_13}. 
We show in Figure \ref{fig:vertexManifold} (a) one example of a regular vertex and in Figure \ref{fig:vertexManifold}(b) and \ref{fig:vertexManifold}(c) two cases where the vertices are not regular and thus they break the manifold property.

In literature \cite{Hoppe13,Litvinov_Lhuillier_13}, reconstruction from sparse data involves the estimation of a mesh which usually partitions the 3D Delaunay Triangulation of a set of sparse points between two sets: free space and matter.
In this case, the sources of non-manifoldness can be induced by tetrahedra intersecting at a vertex, and by tetrahedra with a common edge (non-manifold edge). 
In both cases the non-manifoldness can be fixed, in principle, by cloning and offsetting the vertex of intersection, but two big drawbacks arise: this process would invalidate the Delaunay property, that we need to keep valid in an incremental setting, and, since many non-manifold vertices are generated, rearranging the triangulation and the visibility information for each non-manifold point becomes inefficient. For these two reasons we need a different approach to enforce the manifoldness.

We reconstruct a first manifold with the algorithm proposed in \cite{Litvinov_Lhuillier_13}, slightly modified by a weighting scheme as in \cite{Romanoni15b,Romanoni15a}, bootstrapping from the camera poses, the reconstructed points and the visibility estimated by a Structure from Motion algorithm\footnote{In our implementation we use VisualSFM \cite{wu2011visualsfm}.}. 
The manifold is reconstructed such that it partitions the 3D triangulation of the SfM points between the set $O$ of \emph{outside} tetrahedra, i.e., the subset of the free space outside the manifold (not all the free space tetrahedra will be part of the space outside the manifold), and the complementary set $I$ of inside tetrahedra (i.e., the remaining tetrahedra that represent the matter together with the free space tetrahedra which would invalidate the manifold property). A weight is associated to each tetrahedron and it keeps track of the visibility information, i.e., the camera-to-point viewing rays; in the following, a tetrahedron belongs to free space if its weight is higher than a threshold $t_w$ (in our case $t_w = 4.0$).

The manifold is obtained by three steps. (1)  \emph{Point Insertion}: build the 3D Delaunay triangulation of the 3D SfM points. (2) \emph{Ray tracing and tetrahedra weighting}: for each camera-to-point viewing ray, add a weight $w_1 = 4.0$ to the intersected tetrahedra and a weight $w_2=0.5$ to their neighboring tetrahedra.
Such weighting scheme acts as a smoother of the visibility and avoids the creation of visual artifacts (see \cite{litvinov_Lhiuller14} for a detailed discussion about visual artifacts). (3) \emph{Growing}: initialize a queue $Q$ starting from the tetrahedron with the higher weight. 
Then: (a) pick  the tetrahedron with highest weight from $Q$ and add it to $O$ only if the resulting surface between $O$ and $I$ remains manifold; (b) add the neighboring tetrahedra to the queue $Q$, otherwise discard it; (c) continue iteratively until $Q$ is empty.

\begin{figure}[tp]
  \centering
  \includegraphics[height=0.3\textwidth]{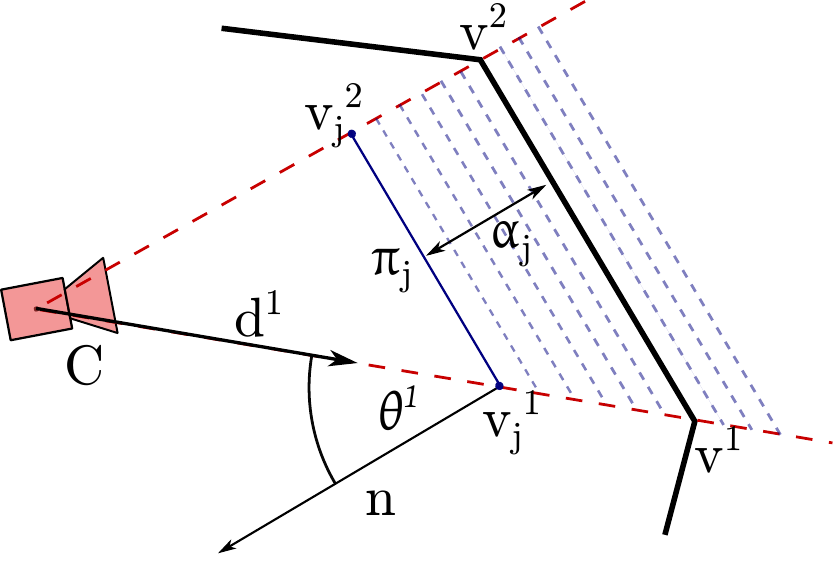}
  \caption{Example of manifold sweeping with respect to the camera $C$}
  \label{fig:sweep}
\end{figure}

\subsection{Mesh Sweeping}
\label{sec:sweep}
After each manifold reconstruction step, we apply the novel mesh sweeping algorithm to look for new 3D points in the neighborhood of the manifold mesh, which we assume to be close to the true model of the scene.
These points will be added to the reconstruction at the next manifold reconstruction iteration.

For each camera, we sweep the visible part of mesh $\mathcal{M}$ along the viewpoint direction (see Figure \ref{fig:sweep}). 
Given a camera located in $C$, we sweep each visible facet $f$ of $\mathcal{M}$ by an amount multiple of $\alpha$ (in our case $\alpha = 3$cm).
Let $v^1$, $v^2$ and $v^3$ be the vertices of $f$, and let $\theta^i$ be the angle between the normal $n$ of the facet and the ray $d^i$ from the camera center to the vertex $v^i$. 
The new vertex $v_j^i$ of the swept facet is now computed as:
$
v_j^i = v^i + \alpha_j \cdot \cos (\theta^i) \cdot d^i,
$
where $\alpha_j = \alpha \cdot k_j$, with $k_j \in \mathbb{N}$ and $-\frac{N}{2}< k_j < \frac{N}{2}$, is the distance swept (let note that is fixed for all the facets). In our case we fixed $N = 20$.

Since all the visible vertices of each triangle are swept by the same amount along the camera-to-point direction, no self-intersections among facets has been induced.

%

\begin{figure}[t]
  \centering
  \includegraphics[height=0.3\textwidth]{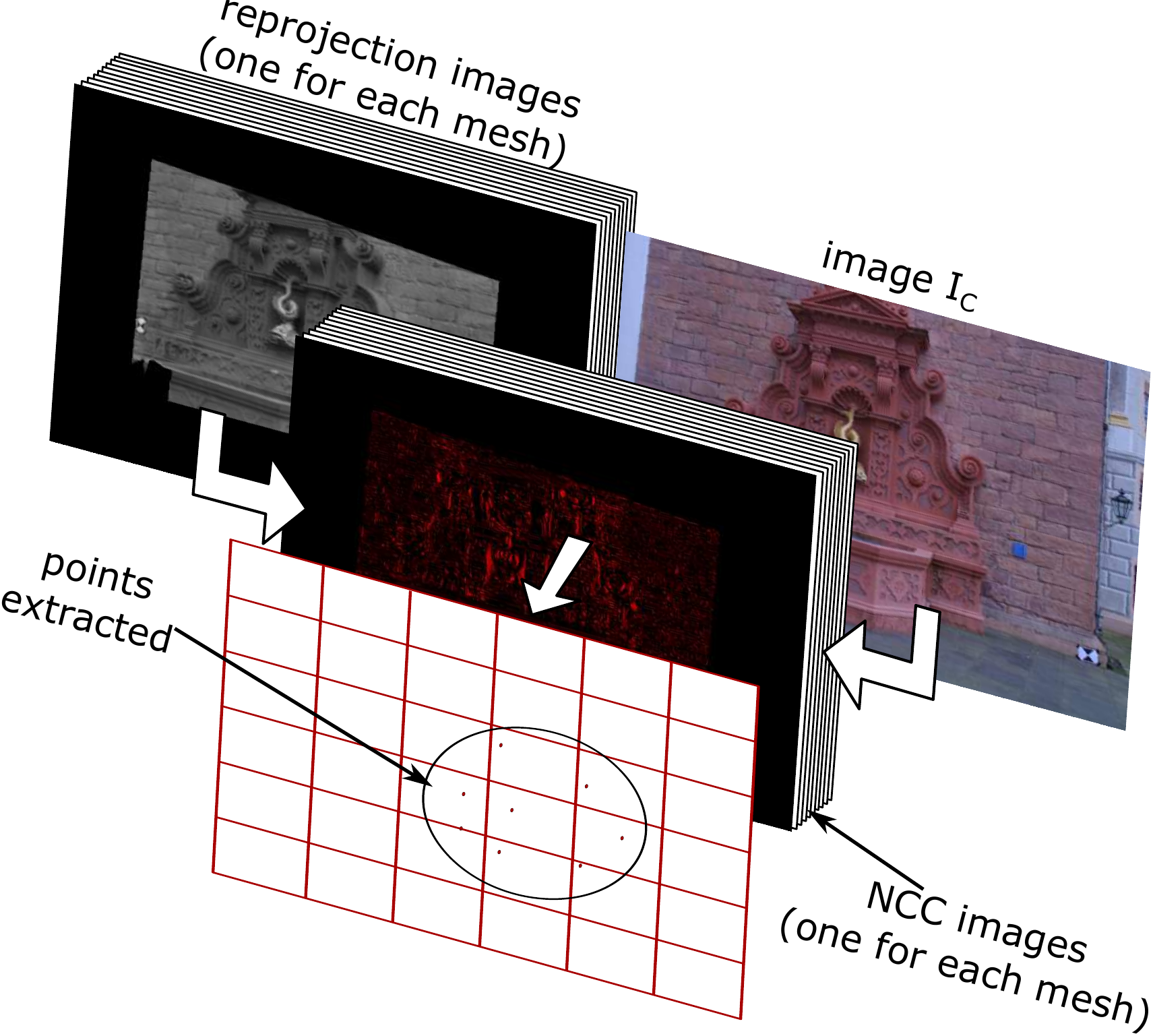}
\caption{3D points extraction process after the mesh sweeping.}
  \label{fig:matching}
\end{figure}

\begin{figure*}[th]
\setlength{\tabcolsep}{1px}
\centering
\begin{tabular}{ccccc}
\includegraphics[height=0.18\textwidth]{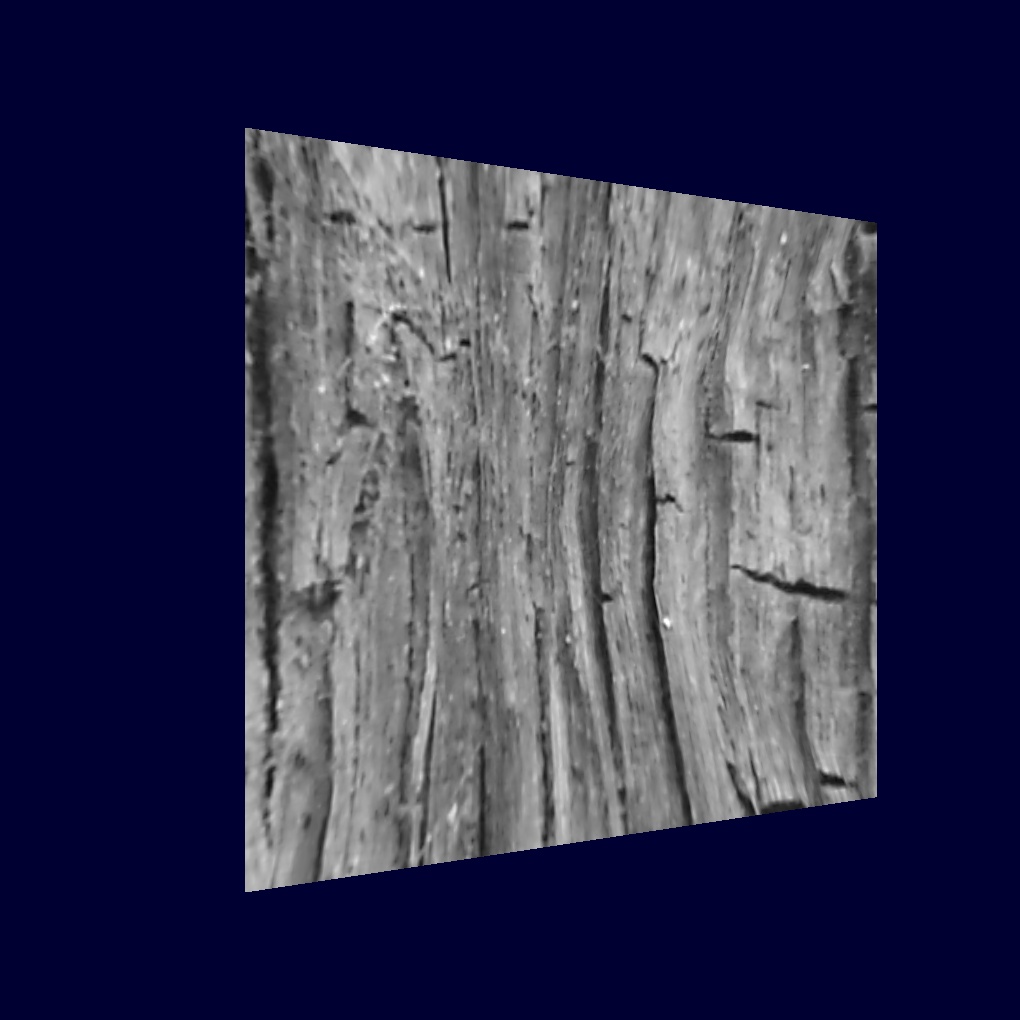}&
\includegraphics[height=0.18\textwidth]{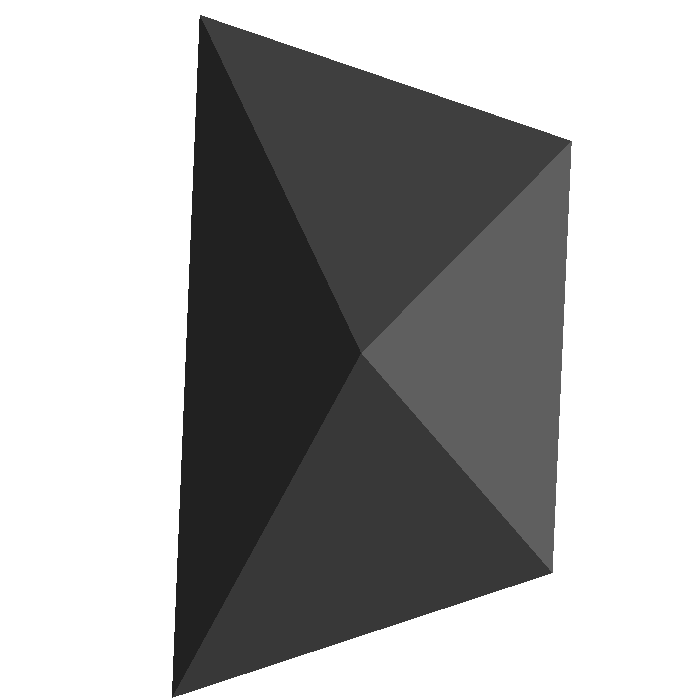}&
\includegraphics[height=0.18\textwidth]{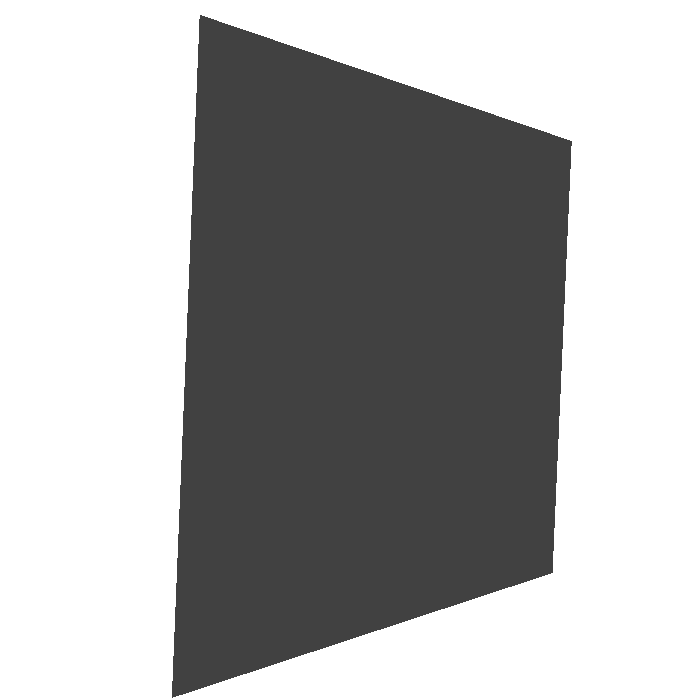}&
\includegraphics[height=0.18\textwidth]{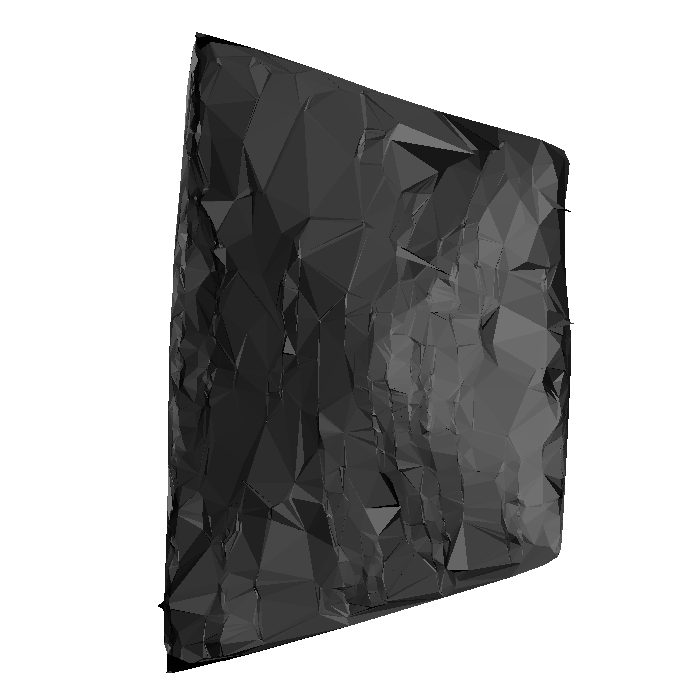}&
\includegraphics[height=0.18\textwidth]{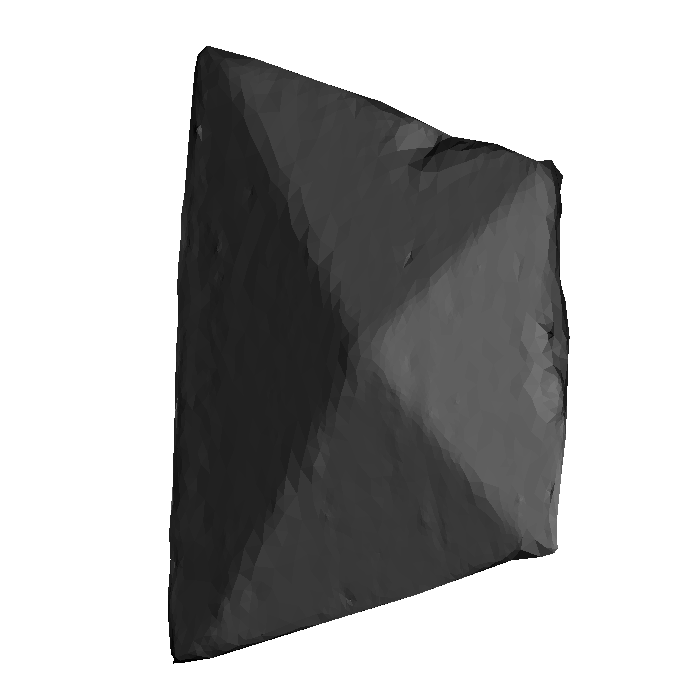}\\
\includegraphics[height=0.18\textwidth]{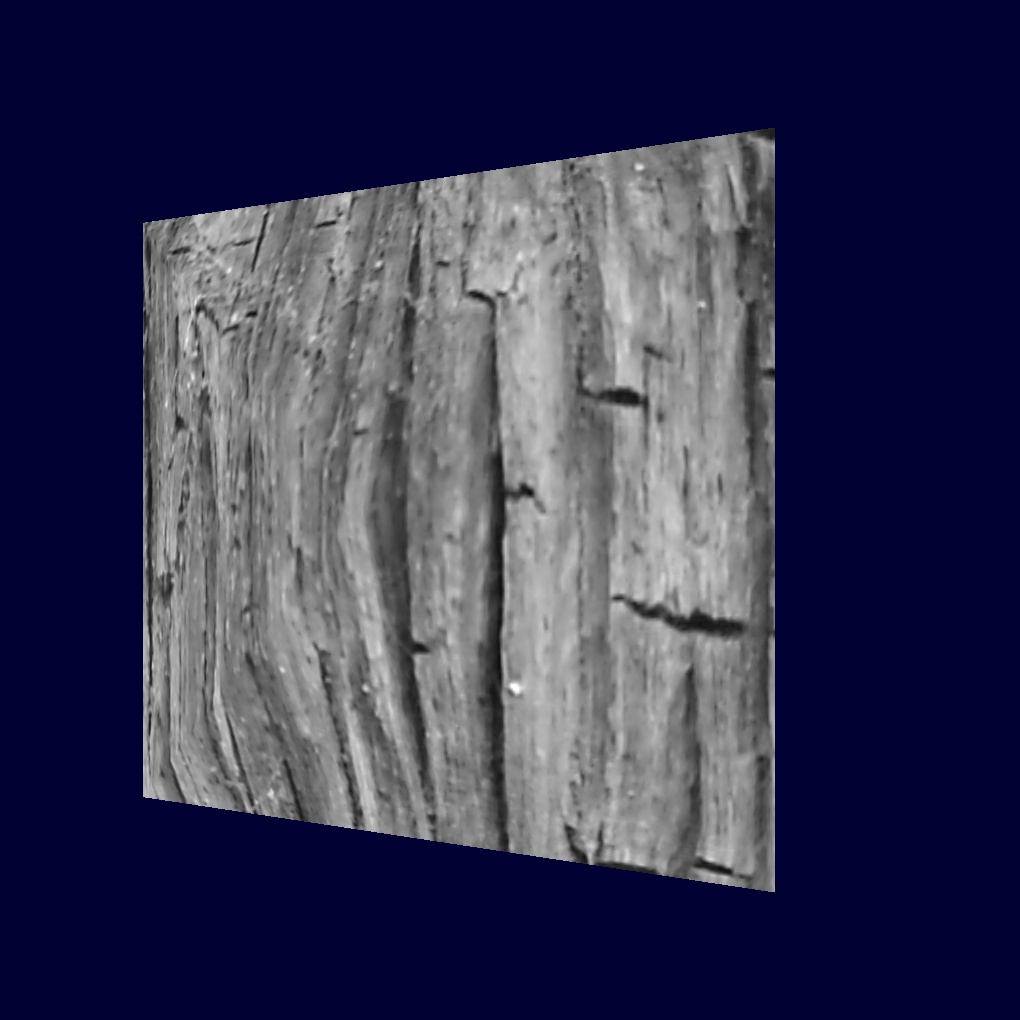}&
\includegraphics[height=0.18\textwidth]{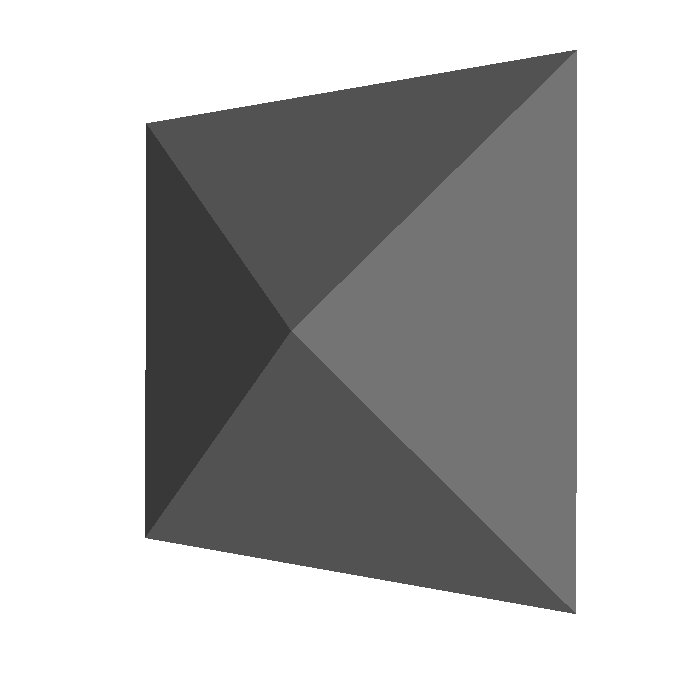}&
\includegraphics[height=0.18\textwidth]{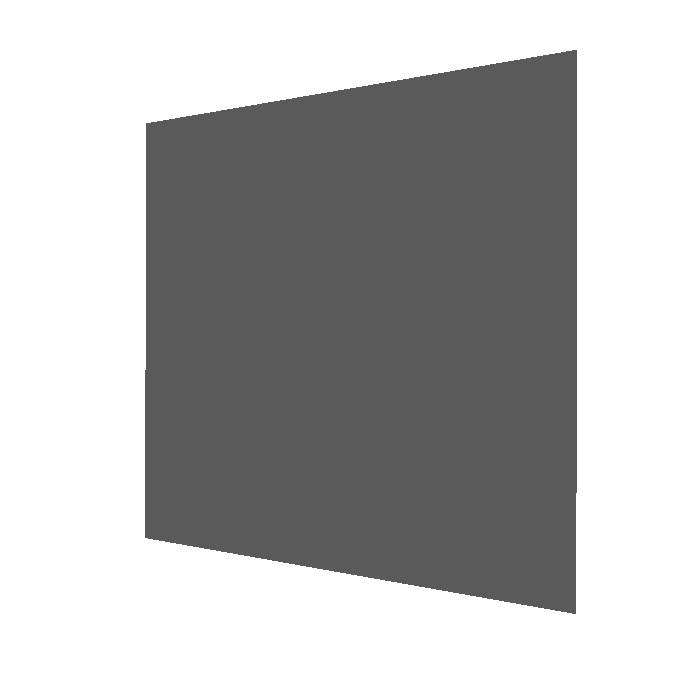}&
\includegraphics[height=0.18\textwidth]{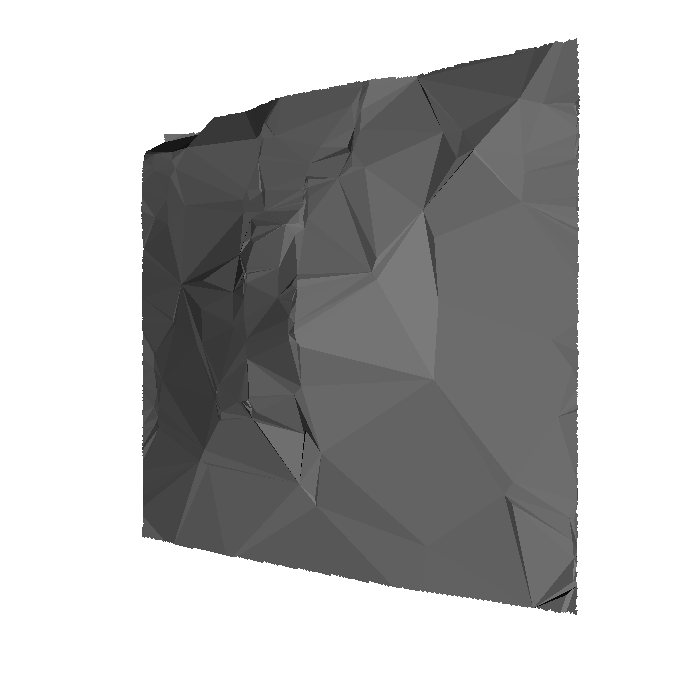}&
\includegraphics[height=0.18\textwidth]{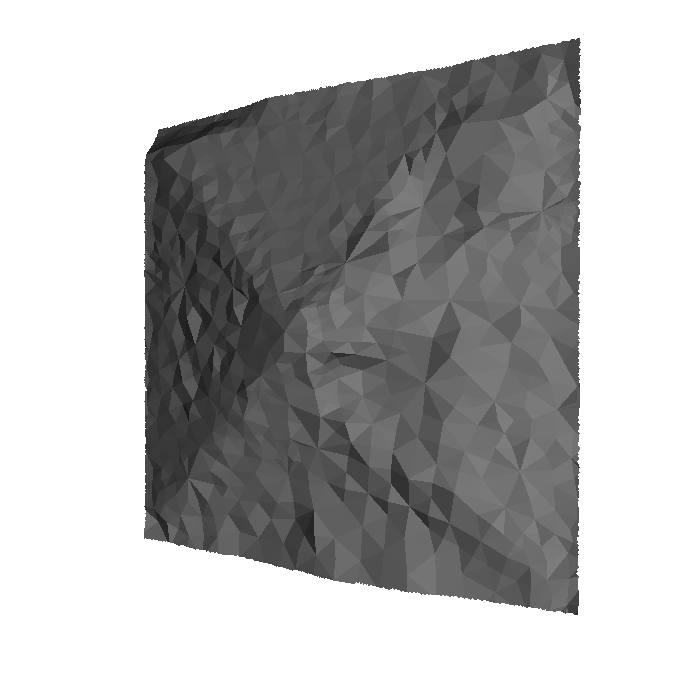}\\
(a) &
(b)&
(c) &
(d) &
(e)\\
\end{tabular}
\caption{Simulated dataset results: (a) textured image, (b) ground-truth, (c) initial mesh, (d) after mesh sweeping, (e) after photometric refinement.}
\label{fig:simulated1}
\end{figure*}

The sweeping process produces a set of $N + 1$ meshes $\mathbb{M} = \{\mathcal{M}_1, \cdots, \mathcal{M}_j ,\cdots, \mathcal{M}_{N + 1}\}$.
For each mesh $\mathcal{M}_j$ we perform pairwise stereo matching between each camera $C$ and the two nearest cameras.
Let $I_C$ be the image seen by $C$, and $I_k$ the image seen by the camera $C_k$, that is one of the two cameras nearest to $C$, therefore $k \in \{1,2\}$. 
For each camera $k$ and mesh $j$, we compute $I_k^j$ as the reprojection of image $I_k$ through the mesh $\mathcal{M}_j$ into camera $C$.
Then, we compute  the Normalized Cross Correlation (NCC) image $I_C$ and the reprojection $I_k^j$. 
The NCC is computed pixel-by-pixel and is weighted by a Gaussian kernel as in \cite{pons2007multi}  (the $\sigma$ for the Gaussian kernel is $\sigma = 8 px$).
The last step aims at collecting the new 3D points. We create a $n_r$x$n_c$ grid over the $N$ images of pixel-by-pixel NCCs computed for each mesh $\mathcal{M}_j$. We collect in each tile the image point with the best NCC above a threshold $t_{\text{NCC}} = 0.98$ among all the $N$ images; therefore, we obtain at most $n_r$x$n_c$ points.
For each collected point, we retrieve its 3D position from the corresponding mesh that generated it.

We are able to perform the whole process image-based, i.e., we act on the image and not on the mesh, so that the most complex computations become independent from the size of the mesh, and they are self-adaptive with respect to the  image resolution. This also allows our approach to be adaptive in term of mesh resolution output.

%

%
Mesh sweeping requires some parameters. $N$ and $\alpha$ need to be fixed in order to keep the sweeping local and, at the same time, to induce an appreciable sweeping of the mesh w.r.t. the image resolution. The values of $N$ or $\alpha$ are also related to the observed scene and the image resolution, but the sweeping algorithm is quite robust against these two parameters.
On the other hand, $\sigma$, $t_{NCC}$, $n_r$ and $n_c$ concur to determine the trade off between precision and recall of the newly estimated 3D points. The threshold $t_{NCC}$ needs to be high so to avoid outliers, even if a good choice of the tile sizes, e.g., 100x100px, manages to make the algorithm robust to outliers and to small variations of parameter setting. Finally, the value of $\sigma$ is usually related the the image resolution, but in our experiments, with different resolutions, we found that the same value of 8px is a good compromise between cross-correlation sensitivity and smoothness.

\subsection{Incremental refinement of the manifold}
To refine the initial mesh, we iterate between the mesh sweeping algorithm, which outputs a set $P$ of new 3D points, and the incremental reconstruction (described in this section)  until convergence is reached (i.e., no more points are available to be added or the number of iteration is greater than $it_{max} = 15$).

The insertion of a point $p\in P$  into the Delaunay triangulation causes the removal of a set $D$ of tetrahedra breaking the Delaunay property. The surface between $O \setminus D$ and $I \cup D$ is not guaranteed to be manifold anymore. 
In order to avoid this, as the authors in \cite{Litvinov_Lhuillier_13}, we define a list of tetrahedra $E \supset D$ and apply the \emph{Shrinking} procedure, i.e., the inverse of Growing:  we subtract iteratively from $O$ only the tetrahedra  $\Delta \in E$ keeping the manifoldness valid.
After this process, it is likely that $D \cap O = \emptyset$.
Whenever $D \cap O \neq \emptyset$ the point $p$ is not added to the triangulation and it is discarded.
Once all points in $P$ have been processed, the queue $Q$ is initialized with the tetrahedra $\Delta \in T \setminus O$ such that  $\Delta \cap \delta O \neq \emptyset$, and the Growing process starts as explained in Section \ref{subsec:manifold}.

\section{Experimental evaluation}
\label{sec:experimental}
The proposed algorithm provides an automatic initialization for mesh-based dense MVS algorithms.
We implemented our approach in C++ using the CGAL library \cite{cgal} for all the computations involving the Delaunay Triangulation and manifold extraction. This allows us to exploit efficient data structures and algorithms for the initial manifold reconstruction tasks.
Conversely, the proposed mesh sweeping algorithm exploits the power of GPU computing; we implemented the main steps with the GLSL shading language of OpenGL \cite{opengl}: a geometry shader computes the normal; the vertex shaders sweep the mesh along the camera viewing rays; the fragment shaders implement the reprojection from one image to an another through the mesh surface, the NCC computation and the thresholding.

\begin{figure}[t]
\setlength{\tabcolsep}{1px}
\centering
\begin{tabular}{c}
\includegraphics[width=0.42\textwidth]{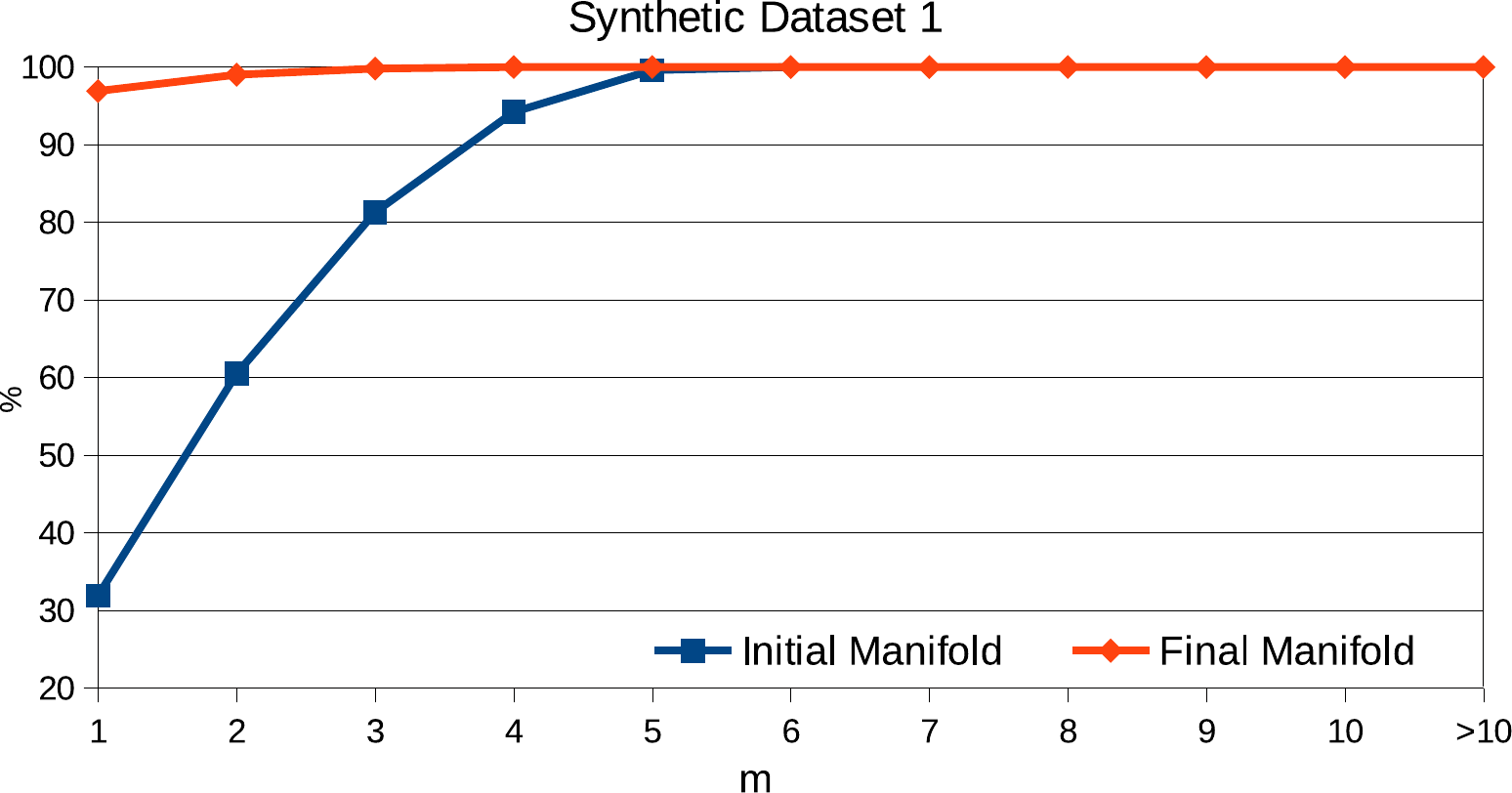}\\
\includegraphics[width=0.42\textwidth]{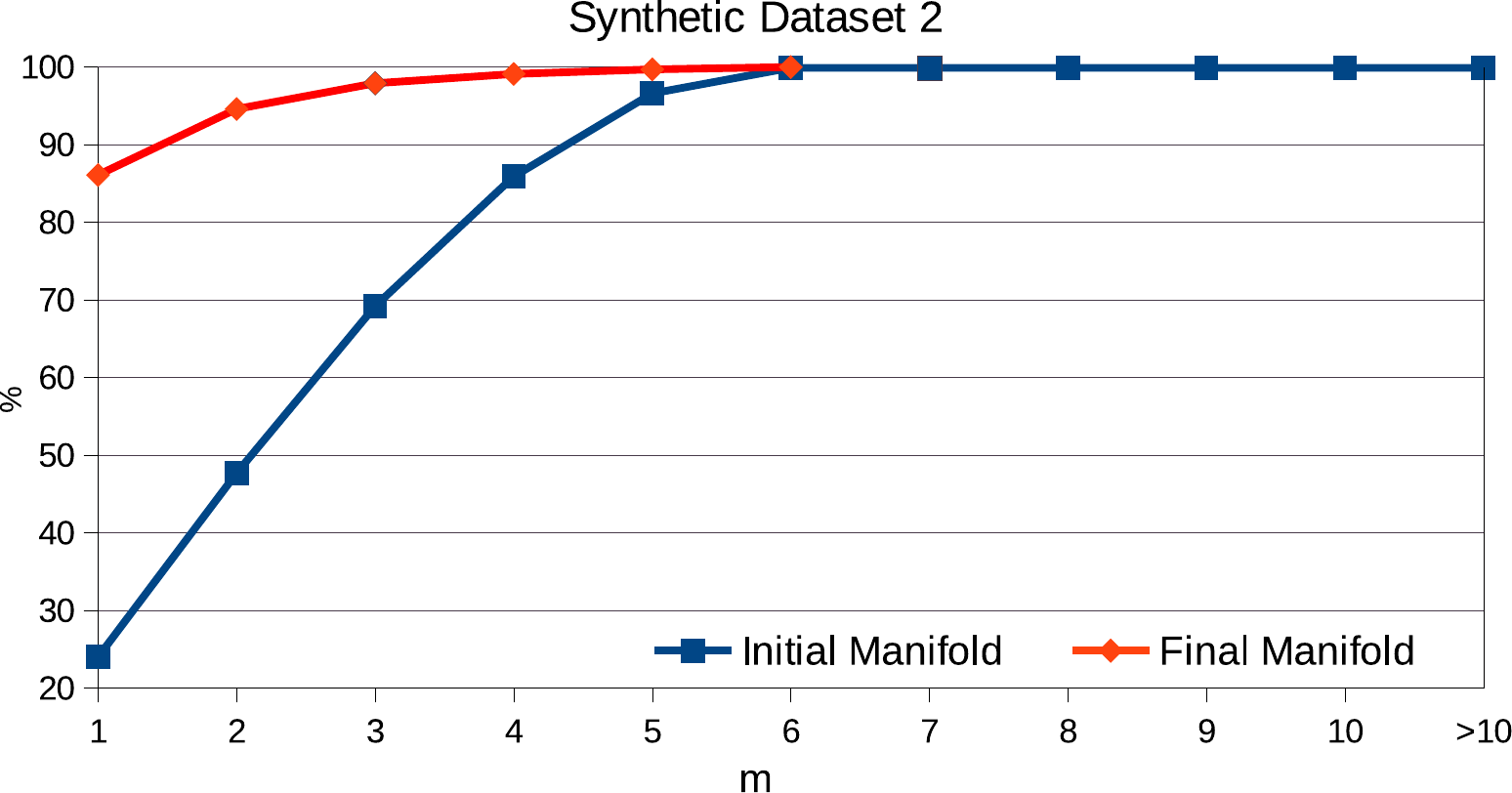}\\
\end{tabular}
\caption{Cumulative distributions of error in the synthetic datasets.}
\label{fig:simulatedhist}
\end{figure}


We performed experiments on both synthetic and real datasets on a 4 Core i7-920 CPU at 2.6Ghz (8M Cache), with 24GB of DDR3 SDRAM and NVIDIA TitanX. Synthetic experiments aim at showing the effectiveness of the mesh sweeping approach to refine a rough manifold.
The experiments on real data show the scalability of the proposed method and its effectiveness compared to CMVS \cite{Fu10}, used as automatic initialization in mesh-based algorithms \cite{pan2015automatic,li2015detail} together with Poisson reconstruction \cite{kazhdan2006poisson}.

In Figure \ref{fig:simulated1} we provide quantitative evaluation for the two synthetic datasets.
The ground-truth meshes we used to generate the image are two pyramids without the squared base facet (Figure \ref{fig:simulated1}(b)), pointing downward for the first dataset, and upward in the second one; the dimensions of these pyramid are 2x2m of squared base and 0.3m height.
The initial point cloud in both cases is made up by the four vertices of the base. 
We bootstraps from only these four points, to show the impact of the proposed approach even if few sparse points are available.
In both cases the initial manifold extracted is a square corresponding to the base of the pyramid (Figure \ref{fig:simulated1}(c)).
The final reconstructions are shown in Figure \ref{fig:simulated1}(d) even if some little artifacts are created due to the noise in the matching process, the extracted surface is close to the ground-truth.
We computed the distance of the meshes with respect to the ground truth by comparing the depth of each pixel from the first camera, similarly to the comparison method proposed in \cite{strecha2008}. 
In Table \ref{tab:resSim} we compare the accuracy of the initial mesh with the accuracy after the proposed mesh sweeping: our method improves significantly the initial mesh; this result is also supported by the improvement on the error distributions in  Figure \ref{fig:simulatedhist}, where the errors are reported as a percentage of depths such that their error $e$ w.r.t. the ground truth is $e<m*Sigma$ (Sigma = 0.06m).
In Table \ref{tab:resSimImp} we applied the photometric refinement of \cite{vu_et_al_2012} to the two meshes and we show the results: the convergence of the refinement after the mesh sweeping is faster and the the reconstruction more accurate.

\begin{figure}[t]
\centering
\includegraphics[width=0.42\textwidth]{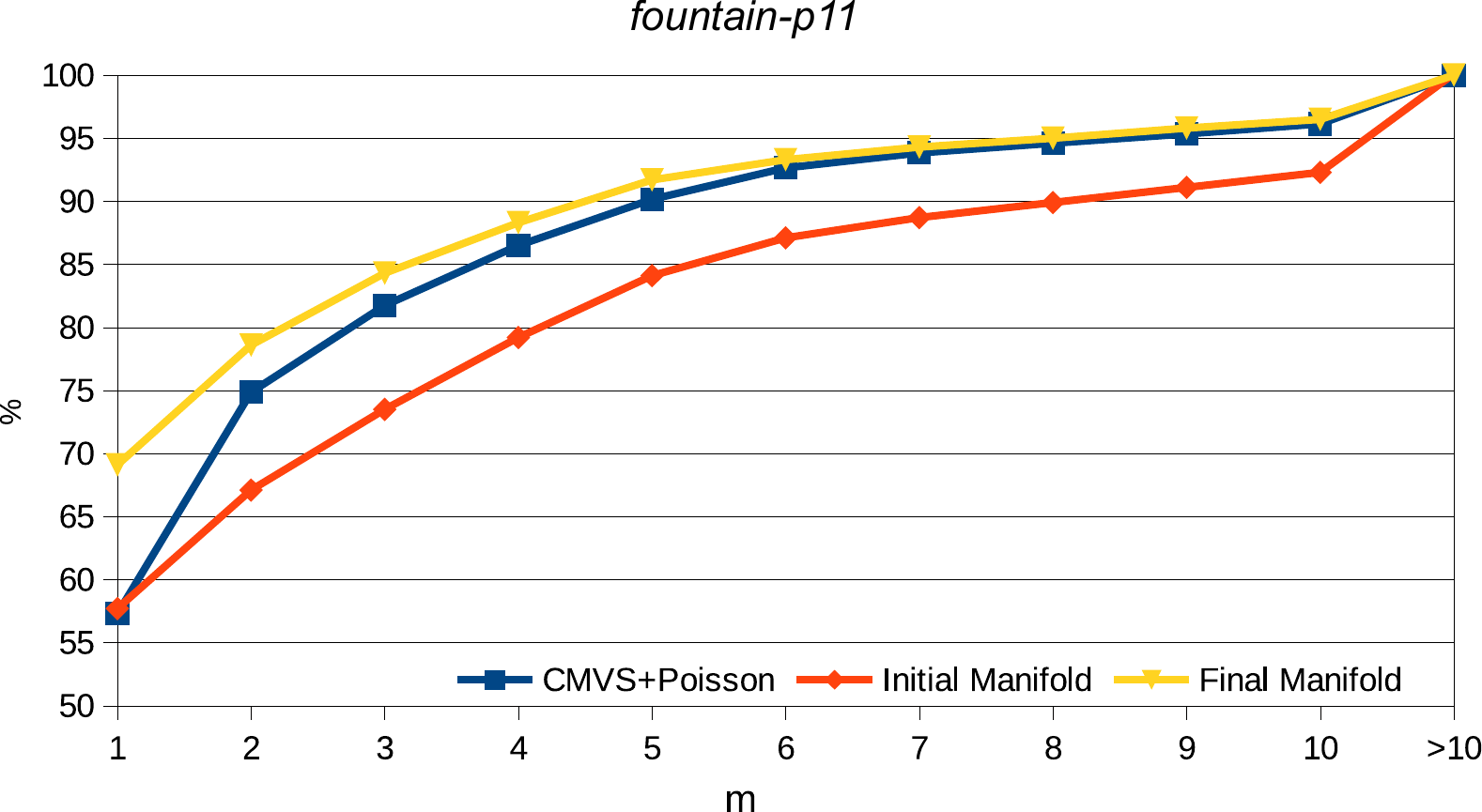}
\caption{Cumulative distribution of errors in the fountain-P11 dataset.}
\label{fig:fountainhist}
\end{figure}



\begin{table}[t]
\caption{Results of the proposed mesh sweeping in the synthetic datasets. All errors are in meters.}
\label{tab:resSim}
\scriptsize
\centering
\begin{tabular}{l|ccc|ccc}
\hline
&\multicolumn{3}{c}{Downward Pyramid}&\multicolumn{3}{c}{Upward Pyramid}\\
              & MEA & MRE & RMS  &  MEA& MRE & RMS  \\
\hline
w/o mesh sweeping  & 0.108 & 0.072 & 0.131 &  0.135 & 0.114 & 0.159 \\
w/ mesh sweeping    & 0.013 & 0.008 & 0.025 &  0.028 & 0.022 & 0.049 \\
improvement   & 88 \% & 89 \% & 81 \% &  79 \% & 81 \% & 69 \% \\
\end{tabular}
\end{table}

\begin{table}[t]
\caption{Photometric refinement with or without the mesh sweeping  in the synthetic datasets. All errors are in meters.}
\label{tab:resSimImp}
\setlength{\tabcolsep}{4px}
\scriptsize
\centering
\begin{tabular}{lc|ccc|ccc}
\hline
&&\multicolumn{3}{c}{Downward Pyramid}&\multicolumn{3}{c}{Upward Pyramid}\\
Initial mesh & n. iter & MEA & MRE & RMS  &  MEA& MRE & RMS  \\
\hline
w/o mesh sweeping &500 & 0.089 & 0.060 & 0.123 &  0.124 & 0.106 & 0.161 \\
w/o mesh sweeping & 5000 & 0.013 & 0.008 & 0.025 &  0.119 & 0.103 & 0.160 \\
w/ mesh sweeping &500& \textbf{0.002} & \textbf{0.001} & \textbf{0.008} &  \textbf{0.012} & \textbf{0.009} & \textbf{0.029} \\
\end{tabular}
\end{table}

\begin{table}[t]
\caption{Comparison between CMVS+Poisson, initial mesh and final mesh. Errors are expressed in Mean Absolute Error (MAE),  Mean Relative Error (MRE) and  Root Mean Square error (RMS).}
\label{tab:resFount}
\scriptsize
\centering
\begin{tabular}{lcccc}
\toprule 
 & MEA & MRE & RMS & num. \\
 & (m) & (m) & (m) & vertices \\
\midrule
CMVS \cite{Fu10} + Poisson  & 0.160 & 0.019 & 0.317& 253410\\
initial mesh \cite{Romanoni15b} & 0.163 & 0.096 & 0.330 & 5880\\
final mesh (Proposed)  & \textbf{0.094} & \textbf{0.012} & \textbf{0.210} & 22336\\
\end{tabular}
\end{table}

\begin{table}[t]
\caption{Photometric refinement with or without the mesh sweeping  in the fountain dataset. All errors are in meters.}
\label{tab:resFountPhoto}
\setlength{\tabcolsep}{4px}
\scriptsize
\centering
\begin{tabular}{lccccc}
\toprule 
Initial & & MEA & MRE & RMS & num. \\
mesh & n iter. & (m) & (m) & (m) & vertices \\
\midrule
w/o mesh sweeping &500& 0.130 & 0.157 & 0.304 & ~1M\\
w/ mesh sweeping &500& \textbf{0.069} & \textbf{0.008} & \textbf{0.196} & ~1M\\
\end{tabular}
\end{table}

We tested the proposed approach on the \emph{fountain-P11} dataset provided in \cite{strecha2008}: the ground-truth is available and the evaluation is performed by comparing the depth map generated from the 6th point of view as suggested in  \cite{strecha2008}, such that the results of the comparison are independent from the quality of the camera calibration (a detailed discussion is available in \cite{strecha2008}). We initialized the manifold reconstruction with the sparse point cloud data estimated with VisualSFM \cite{wu2011visualsfm}.

\begin{figure}[t]
\setlength{\tabcolsep}{1px}
\centering
\begin{tabular}{cccc}
\includegraphics[width=0.25\columnwidth,height=0.1\textwidth]{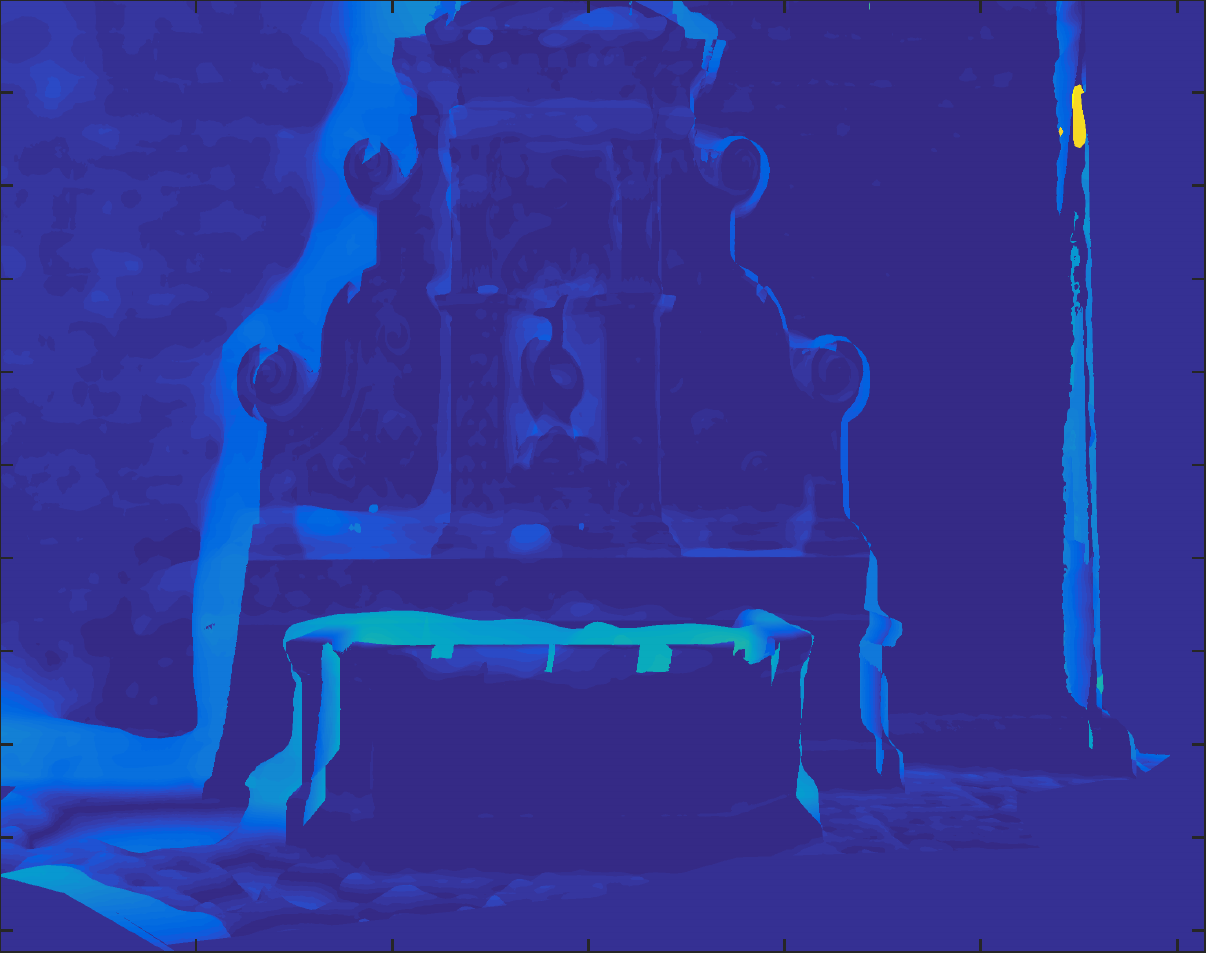}&
\includegraphics[width=0.25\columnwidth,height=0.1\textwidth]{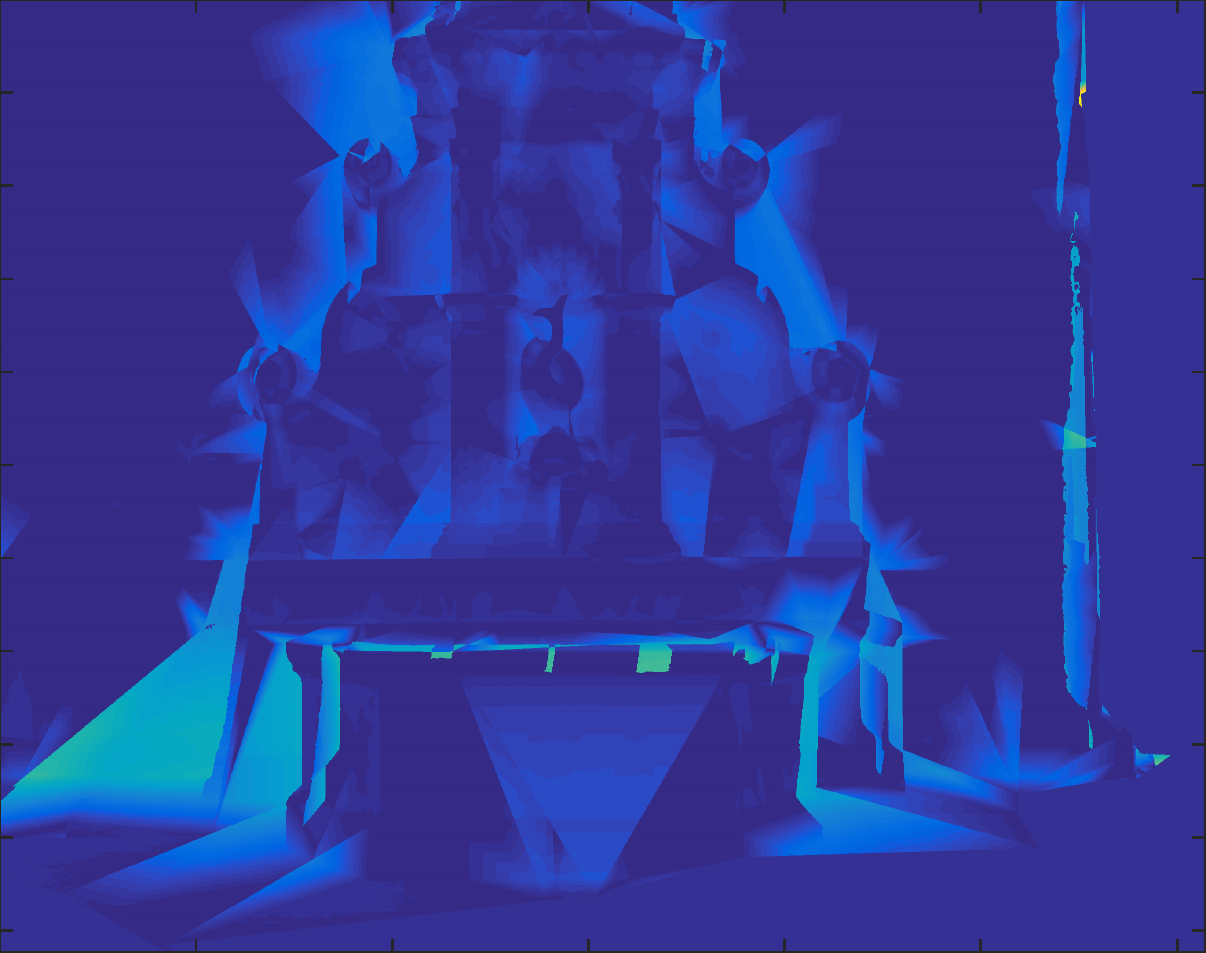}&
\includegraphics[width=0.25\columnwidth,height=0.1\textwidth]{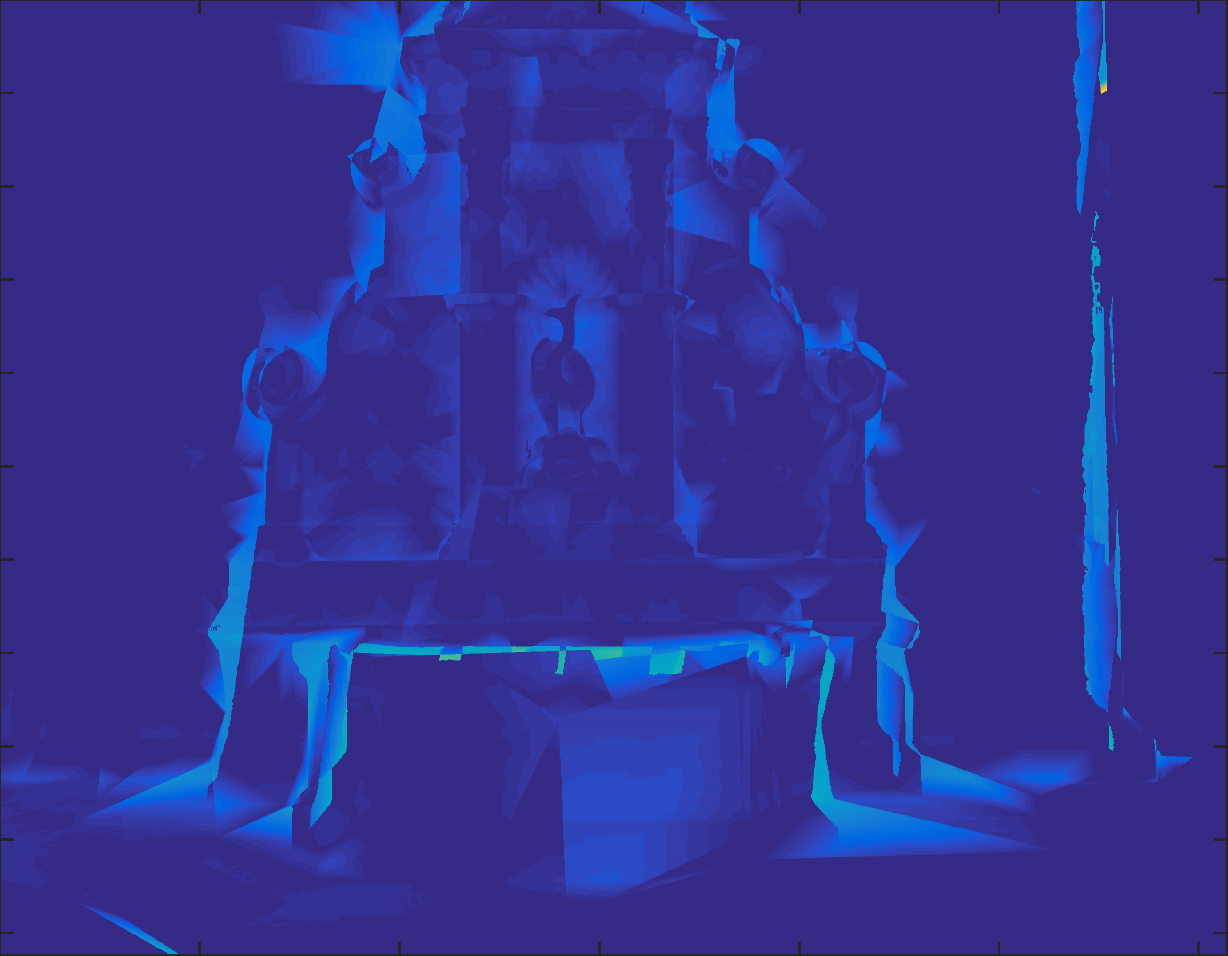}&
\includegraphics[width=0.25\columnwidth,height=0.1\textwidth]{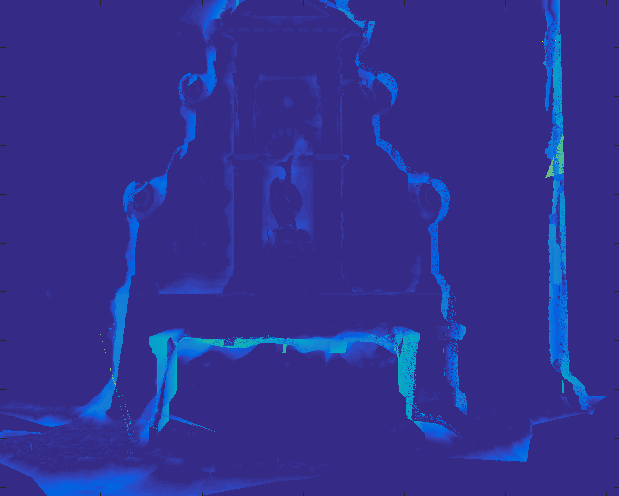}\\
\includegraphics[width=0.25\columnwidth,height=0.1\textwidth]{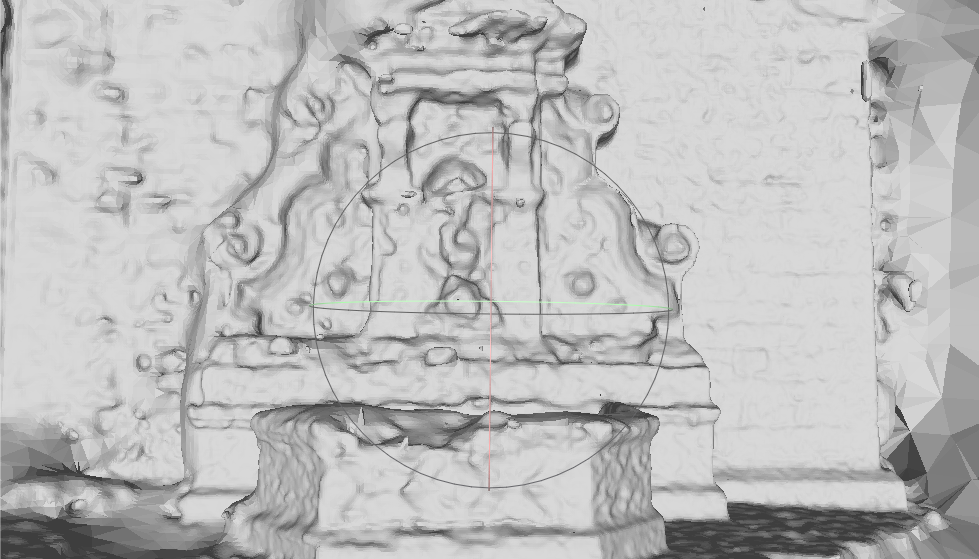}&
\includegraphics[width=0.25\columnwidth,height=0.1\textwidth]{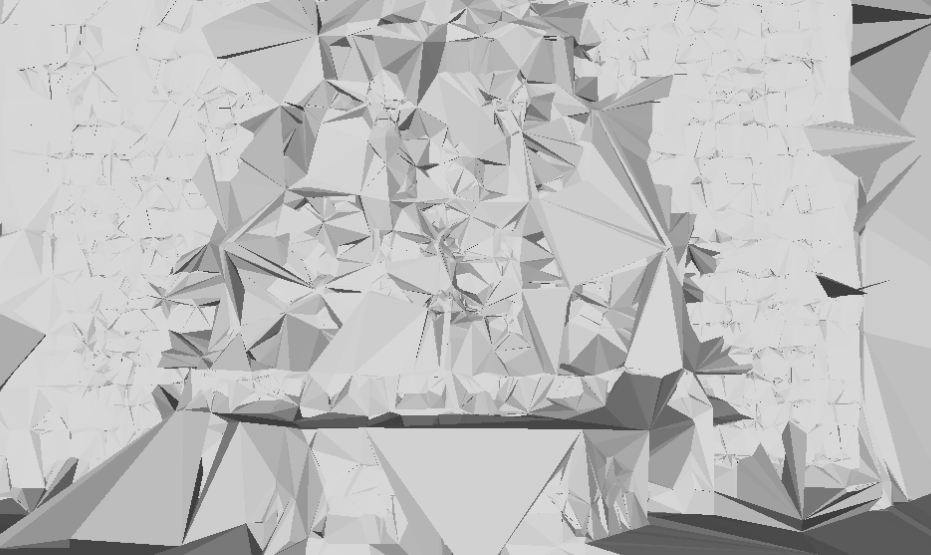}&
\includegraphics[width=0.25\columnwidth,height=0.1\textwidth]{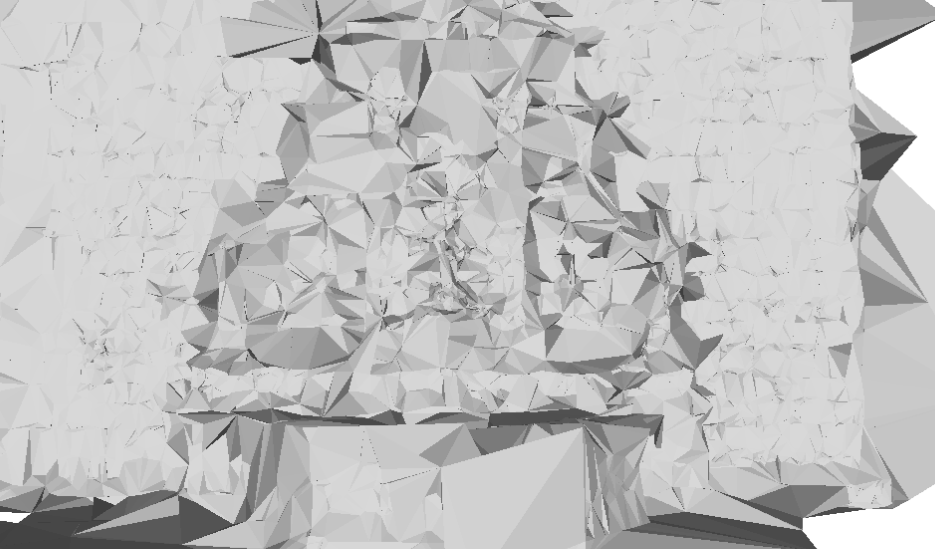}&
\includegraphics[width=0.25\columnwidth,height=0.1\textwidth]{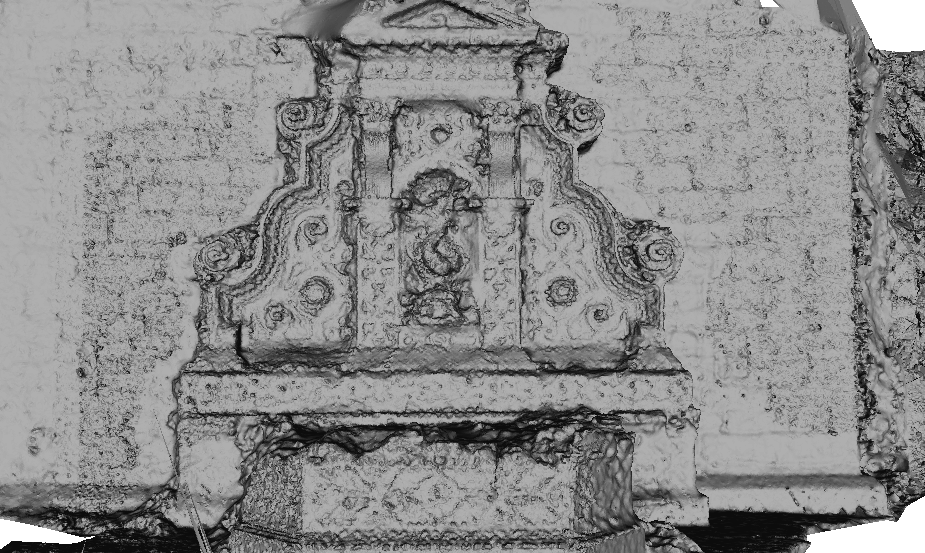}\\
\includegraphics[width=0.25\columnwidth,height=0.1\textwidth]{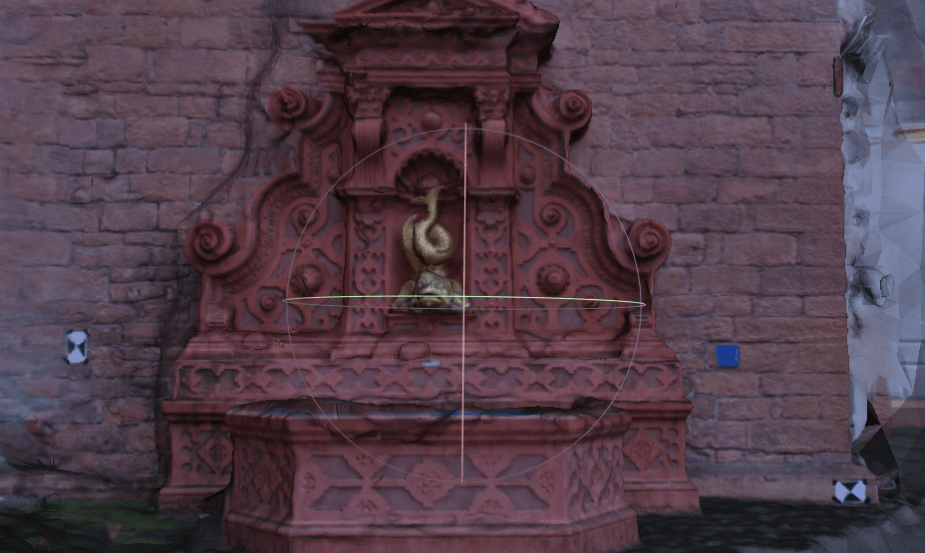}&
\includegraphics[width=0.25\columnwidth,height=0.1\textwidth]{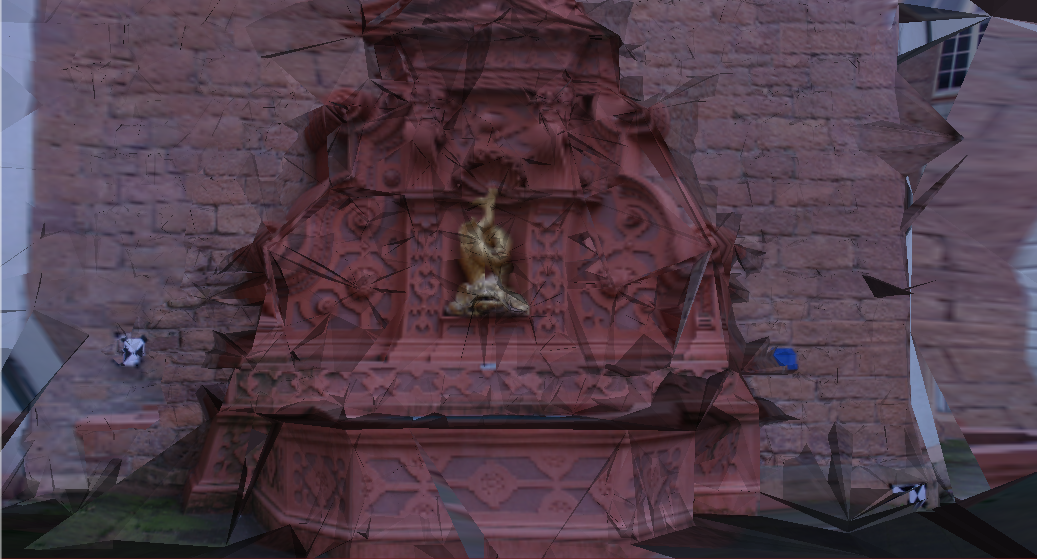}&
\includegraphics[width=0.25\columnwidth,height=0.1\textwidth]{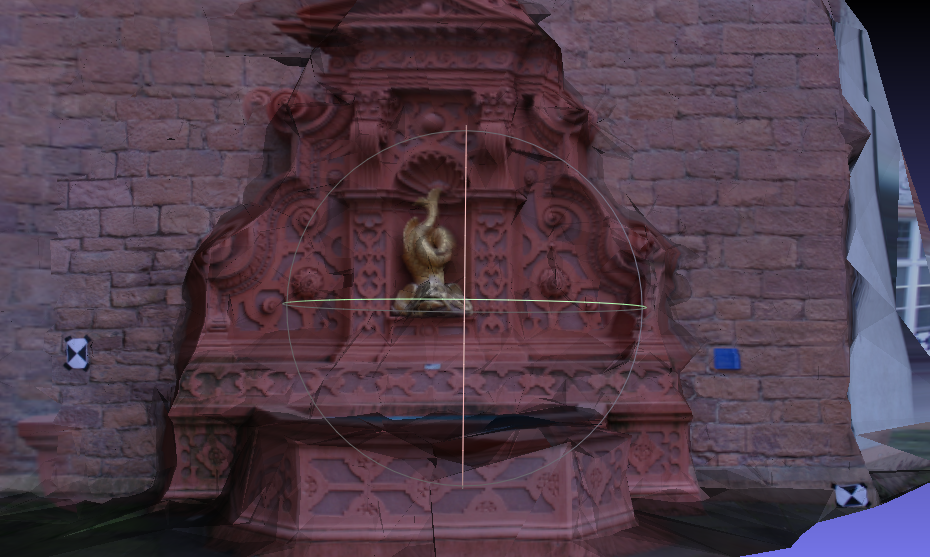}&
\includegraphics[width=0.25\columnwidth,height=0.1\textwidth]{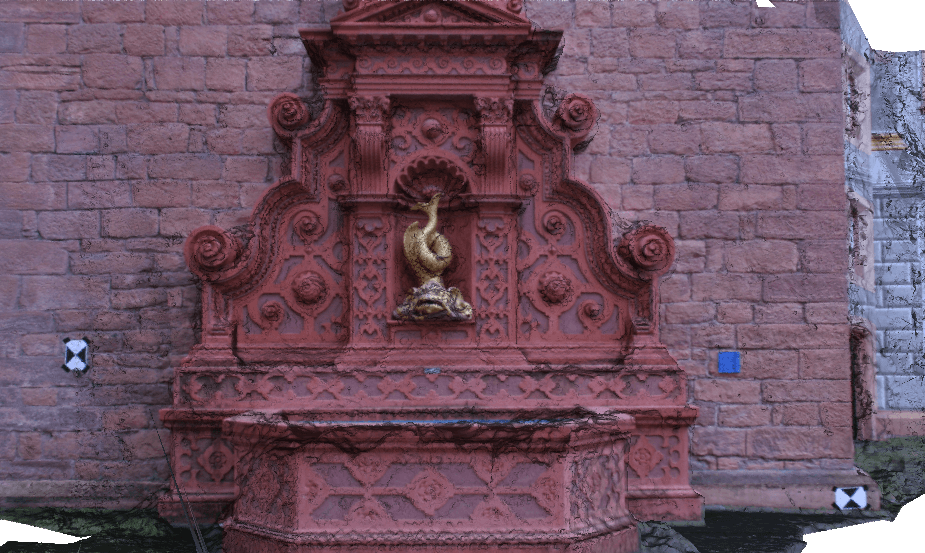}\\
CMVS &
Initial&
After Mesh&
Photometric\\
Poisson &
manifold&
 Sweeping&
Refinement\\
\end{tabular}
\caption{Reconstruction results. First row: depth errors darker blue encodes lower errors. Second row: untextured reconstruction. Third row: textured reconstruction.}
\label{fig:fountainIm}
\end{figure}

Table \ref{tab:resFount} shows the accuracy of the initial manifold, estimated in a similar fashion to \cite{Romanoni15b}, the manifold after the mesh sweeping and the reconstruction performed by CMVS+Poisson, where the octree depth of the Poisson reconstruction is 11 (see \cite{kazhdan2006poisson}). In Figure \ref{fig:fountainhist} we report the cumulative distribution of the errors for each algorithm with the same convention as before.
As Figure \ref{fig:fountainIm} shows, the initial manifold represents accurately the wall, while some parts of the fountain present artifacts, indeed, the accuracy of this initial guess is lower than the outcome of CMVS+Poisson.
The mesh sweeping refinement improves the mesh accuracy, both respect to the initial mesh and to CMVS+Poisson reconstruction, even if the resolution of the latter is significantly higher. Our approach avoids to reconstruct redundant data, differently from the point-based CMVS, which reconstruct as much points as possible, even if they are coplanar.
Table \ref{tab:resFountPhoto} shows that the proposed approach improves the convergence of the variational photometric refinement described in \cite{vu_et_al_2012} leading to a more accurate reconstruction. 
The mean timing per-iteration is 318s for mesh sweeping and 55s for manifold reconstruction: the mesh sweeping is the most demanding task, but thanks to its GPU implementation it would easily take advantage from more powerful hardware.

\begin{figure}[t]
\setlength{\tabcolsep}{1px}
\centering
\begin{tabular}{c}
\includegraphics[width=0.68\columnwidth]{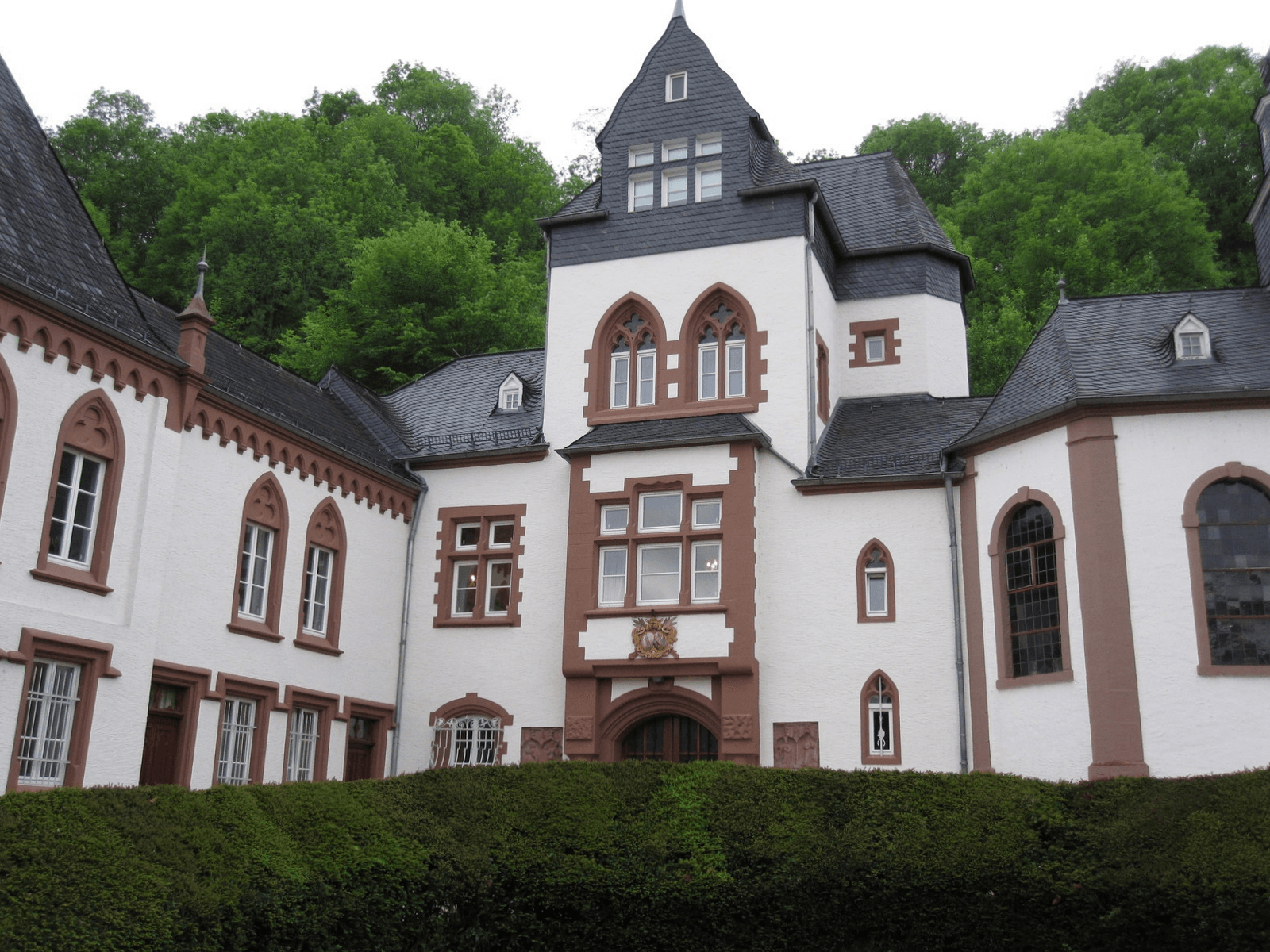}\\
\includegraphics[width=0.68\columnwidth]{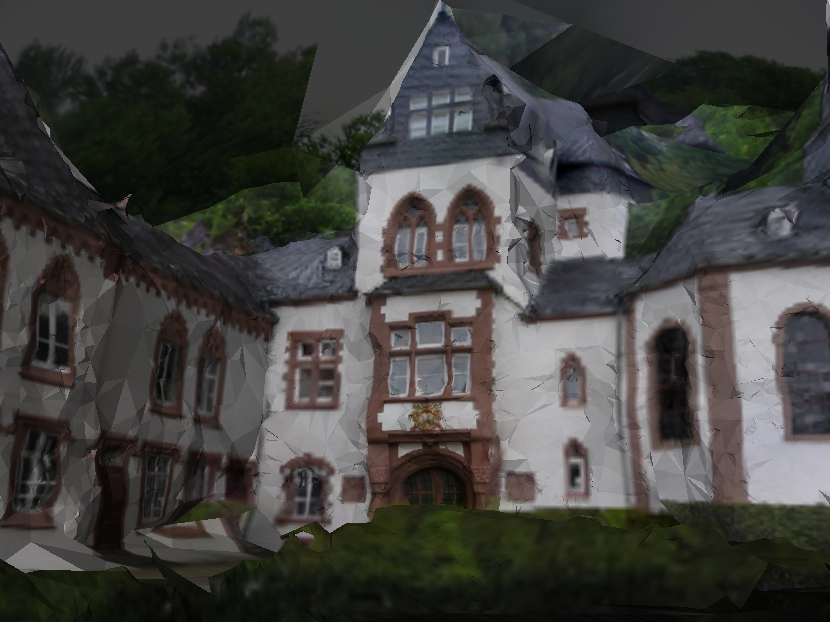}\\
\end{tabular}
\caption{One frame of the Dagstuhl and reconstruction with the proposed method.}
\label{fig:Dagstuhl}
\end{figure}

\begin{figure}[t]
\centering
\includegraphics[width=0.42\textwidth]{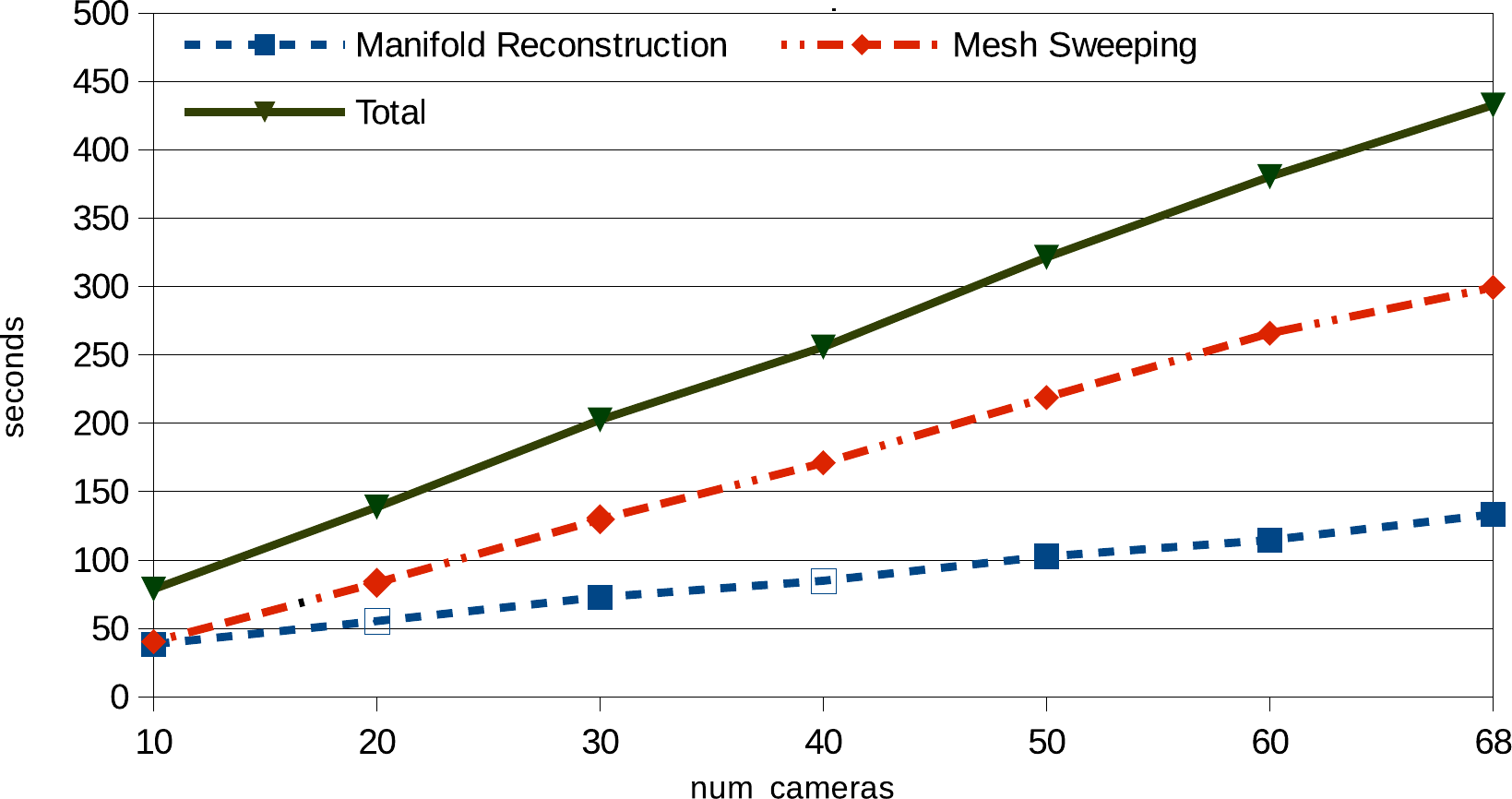}
\caption{Processing time for the Dagstuhl dataset.}
\label{fig:scalability}
\end{figure}

Finally we tested the scalability of the proposed approach with a real dataset, named Dagstuhl, of 68 $1600\text{x}1200$ images (see Figure \ref{fig:Dagstuhl}). 
We tested the algorithm considering different subsets of frames to understand experimentally the dependence with the number of cameras of the proposed algorithm and the two steps (manifold reconstruction and mesh sweeping) separately. 
As Figure \ref{fig:scalability} points out, both steps and the whole algorithm show experimentally a linear relation with the number of cameras; this suggests that our approach is scalable and suitable for large-scene reconstructions.

%

\section{Future Works and Conclusion}
\label{sec:conclusion}
In this paper we proposed a new approach to reconstruct an initialization for mesh-based multi-view stereo algorithm such that the whole pipeline is fully automatic, taking into account that state-of-the-art algorithms are not able to evolve correctly non-manifold meshes.
We build a first rough mesh with an incremental manifold reconstruction algorithm, then, we iterate between the novel mesh sweeping refinement and the manifold reconstruction.
The former sweeps the mesh along the cameras viewing direction and seeks new pairwise stereo matches in these swept meshes; the corresponding 3D points and visibility information are added to the incremental manifold surface algorithm. 
Our approach improves the initial mesh accuracy and resolution, and also with respect to the reconstruction based on CMVS+Poisson, usually adopted as mesh-based MVS initialization.
The presented algorithm has the strong advantage of being suitable for large scale, resolution-adaptive, and incremental 3D reconstruction scenarios, while keeping the manifold properties.

As future work we would improve the robustness with respect to untextured regions, in particular when more than one main untextured plane is present, and we would like to test other stereo matching measures. Moreover we would like to test in our scenarios some ideas from \cite{chauve2010robust}, such as the addition of the ghost primitives.

\section{Acknowledgements}
We thank NVIDIA who kindly supported this work through the Hardware Grant Program.

{\small
\bibliographystyle{ieee}
\bibliography{biblio}
}
\end{document}